\documentclass[journal]{IEEEtran}
\usepackage{cite}

\ifCLASSINFOpdf
  \usepackage[pdftex]{graphicx}
\else
  \usepackage[dvips]{graphicx}
\fi
\usepackage[cmex10]{amsmath}
\usepackage{algorithmic}

\usepackage{array}

\ifCLASSOPTIONcompsoc
  \usepackage[caption=false,font=normalsize,labelfont=sf,textfont=sf]{subfig}
\else
  \usepackage[caption=false,font=footnotesize]{subfig}
\fi
\usepackage{dblfloatfix}

\usepackage{booktabs}

\usepackage{url}

\usepackage{float}
\usepackage{multirow}
\usepackage{enumerate}
\usepackage{amsthm}
\usepackage{nidanfloat}

\newtheoremstyle{mystyle}
  {}
  {}
  {\itshape}
  {}
  {\bfseries}
  {.}
  { }
  {}
\theoremstyle{mystyle}
\newtheorem{definition}{Definition}

\newtheorem{corollary}{Corollary}

\newtheorem{remark}{Remark}

\begin{document}
%
\title{Zero-Bias Deep Learning for Accurate Identification of Internet of Things (IoT) Devices}



\author{Yongxin~Liu,
        Jian~Wang,
        Jianqiang~Li,
        Houbing~Song,~\IEEEmembership{Senior Member,~IEEE,}
        Thomas~Yang,~\IEEEmembership{Senior Member,~IEEE,} 
        Shuteng~Niu and
        Zhong~Ming 
\thanks{Jianqiang Li and Zhong Ming are with the College of Computer Science and Software Engineering, Shenzhen University, China}
\thanks{Yongxin Liu, Jian Wang, Houbing Song, Thomas Yang and Shuteng Niu are with the Department of Electrical Engineering and Computer Science, Embry-Riddle Aeronautical University, Daytona Beach, FL 32114}
\thanks{Corresponding authors: Jianqiang Li and Houbing Song}
\thanks{Manuscript received May 4, 2020; accepted for publication on 21 August 2020. }}

\markboth{IEEE Internet of Things Journal,~Vol.~11, No.~4, December~2020}%
{Shell \MakeLowercase{\textit{et al.}}: Bare Demo of IEEEtran.cls for Journals}
%



\IEEEtitleabstractindextext{%
\begin{abstract}
The Internet of Things (IoT) provides applications and services that would otherwise not be possible. However, the open nature of IoT make it vulnerable to cybersecurity threats. Especially, identity spoofing attacks, where an adversary passively listens to existing radio communications and then mimic the identity of legitimate devices to conduct malicious activities. Existing solutions employ cryptographic signatures to verify the trustworthiness of received information. In prevalent IoT, secret keys for cryptography can potentially be disclosed and disable the verification mechanism. Non-cryptographic device verification is needed to ensure trustworthy IoT. In this paper, we propose an enhanced deep learning framework for IoT device identification using physical layer signals. Specifically, we enable our framework to report unseen IoT devices and introduce the zero-bias layer to deep neural networks to increase robustness and interpretability. We have evaluated the effectiveness of the proposed framework using real data from ADS-B (Automatic Dependent Surveillance-Broadcast), an application of IoT in aviation. The proposed framework has the potential to be applied to accurate identification of IoT devices in a variety of IoT applications and services. Codes and data are available in \cite{gt9v-kz32-20}.
\end{abstract}

}

\maketitle

\IEEEdisplaynontitleabstractindextext

%
\IEEEpeerreviewmaketitle

\section{Introduction}
%
%
%
%

The Internet of Things (IoT) is characterized by the interconnection and interaction of smart objects (objects or devices with embedded sensors, onboard data processing capability, and a means of communication) to provide applications and services that would otherwise not be possible \cite{IIOT17}. The convergence of sensor, actuator, information, and communication technologies in IoT produces massive amounts of data that need to be sifted through to facilitate reasonably accurate decision-making and control \cite{BDA19}. Big data analytics has the potential to enable the move from IoT to real-time control \cite{7406686}. However, due to the open nature of IoT, IoT is subject to cybersecurity threats \cite{8897627,liu2019domain}. One typical cybersecurity threat is identity spoofing attacks where an adversary passively collect information and then mimic the identity of legitimate devices to send fake information or conduct other malicious activities. Such attacks can be extremely dangerous when appear in critical infrastructures \cite{skarmeta2014decentralized}.

Conventional approaches to prevent identity spoofing attacks employ cryptographic algorithms to verify that a trusted source generates a message. However, the cryptographic approaches depend on the secrecy of encryption keys and encounter challenges from the open and heterogeneous ecosystems of IoT. For example, a number of commercially successful IoT systems, which do not operate with cryptographic keys, require a huge investment to become cryptographically secure \cite{wang2018fountain}. Therefore, there is a need for non-cryptographic solutions to verify the identify of IoT devices, thus ensuring trustworthy IoT. 

Non-cryptographic IoT device identification is inspired by signal identification technology in speech and acoustic signal processing \cite{yue2018software}. The assumption is that each each signal source modulate its unique features into the propagated signals. Comparably, in non-cryptographic IoT device identification, we assume that each wireless transmitter randomly picks up certain types of imperfectness (a.k.a, radiometric fingerprint) during their manufacture \cite{wang2016wireless} and could be reflected in the demodulated signals. Existing works on non-cryptographic device identification can be classified into two categories: specific feature recognition and deep learning. Specific feature-based approaches focus on deriving distinctive features (a.k.a, transmitter fingerprints) from received signals \cite{wang2019integration,zou2017tagfree} to recognize known devices. Deep learning based approaches do not require knowing devices' radiometric characteristics and shows even higher accuracy \cite{chen2019deep,restuccia2019deepradioid}.  However, the challenge of applying deep learning approaches for IoT device identification lies in two aspects: unseen device recognition, and model interpretability. The first challenge requires deep neural networks to report unseen devices rather than erroneously associating them with known ones. The second challenge requires that the behaviors of neural networks to be interpretable.

In this paper, we propose an enhanced deep learning framework for accurate and interpretable identification of IoT devices with mathematically assured performance. We propose a zero-bias dense layer for Deep Neural Networks to jointly verify known devices and identify unknown ones. The effectiveness of the proposed framework in handling massive signal recognition and improving the performance of traditional neural networks has been demonstrated. The contributions of this paper are as follow:
\begin{itemize}
    \item We provide a novel enhancement, the zero-bias layer, to replace the last dense layer in conventional neural networks to increase its interpretability without losing accuracy.
    \item We provide a novel technique to characterize how well a neural network can distinguish from different classes.
    \item We enable our framework to automatically report unknown devices rather than erroneously associating them with known ones.
\end{itemize}

Our research offers not only a solution to accurate identification of IoT devices, thus useful in promoting trustworthy IoT, but also a deep learning framework for intrusion detection. In addition, the introduction of zero-bias layer in deep neural networks represents an advance in deep learning, thus leveraging deep learning to enable the move from IoT to real-time control. 

The remainder of this paper is organized as follows: A literature review of non-cryptographic device identification is presented in Section~\ref{sectRW}. We formulate our problem in Section~\ref{sectPD} with methodology presented in Section~\ref{sectMM}. Performance evaluation is presented in Section~\ref{sectEED} with conclusions in Section~\ref{sectCC}.

\section{Related works}
\label{sectRW}
Non-cryptographic device identification is emerging as a solution to Physical layer security of IoT. Coresponding methods can be classified into two categories: specific feature based and deep learning based. 

\subsection{Specific feature based approaches}
The specific feature based approaches require human efforts to discover distinctive features for device identification. The methods rely on the fact that there are various manufacturing imperfectnesses in wireless devices' RF frontends. These imperfectnesses do not degrade the communication quality but can be exploited to identify each transmitter uniquely. Those features are named Physical Unclonable Features (PUF) \cite{chatterjee2018rf,herder2014physical}). There are two categories of PUFs: error pattern and transient patterns.

In error pattern approaches, it is assumed that the statistical properties of received symbols' noise could uniquely profile wireless devices. In \cite{azarmehr2017wireless}, the authors show that phrase error of Phase Lock Loop in transmitters can provide promising results even with low Signal-to-Noise Ratio (SNR). In \cite{zhuang2018fbsleuth}, the authors use the difference between received signals and theoretical templates to construct error vectors. Error vectors' statistics and time-frequency features are combined as fingerprints for transmitter identification. In \cite{peng2019deep}, the authors employ differential constellation trace figure (DCTF) to capture the time-varying modulation error of Zigbee devices. They then develop their low-overhead classifier to identify 54 Zigbee devices. 

In transient pattern approaches, it is assumed that a malicious entity can not forge the transient response characteristic of wireless transmitters\cite{danev2010attacks}. Transient patterns are commonly seen at the beginning and end of wireless packet transmission. In \cite{polak2015identification},  nonlinear in-band distortion and spectral regrowth of the signals are utilized to distinguish the masquerade emitter. In \cite{kose2019rf}, the authors employ the transient energy spectrum on transmitters' turn-on amplitude envelops to identify, and they show that frequency-domain features outperform time-domain features. 

Feature-based approaches require efforts to manually extract features or high-order statistics for different scenario. Therefore, more effortless and versatile methods are required.

\subsection{Deep neural network based approaches}
Deep Neural Networks (DNNs) are frequently used as a general-purpose BlackBox for pattern recognition. Naturally, they are applied to perform device-specific identification.

A typical DNN enabled wireless device identification system employs convolutional layers to extract latent features. Convolutional layers apply filters (a.k.a., kernels) to obtain helpful information automatically. Such benefit reduces the hardship of manual feature discovery. In \cite{yu2019radio}, the authors provide a novel method that perform the signal denoising and emitter identification simultaneously using an autoencoder and a Convolution Neural Network (CNN). Their solution shows promising results even with low SNR. Similar work in \cite{huang2017communication} employs stacked denoising auto-encoder and show similar results. DNNs perform well even on raw signals. In \cite{riyaz2018deep}, the authors provide an optimized Deep Convolutional Neural Network to classify SDR-based emitters in 802.11AC channels, they show that, even by using raw signals without feature engineering, CNN surpasses the best performance of conventional statistical learning methods. In \cite{morin2019transmitter}, neural networks were trained on raw IQ samples using the open dataset\footnote{\url{https://wiki.cortexlab.fr/doku.php?id=tx-id}} from CorteXlab. Their work also show similar results. Compare with specific feature based approach, deep neural networks dramatically reduce the requirement of domain knowledge and the quality of fingerprints. 

In general, DNNs are becoming a promising building block in non-cryptographic wireless device identification. DNNs encounter a challenge in terms of anomaly detection, which requires that deep learning enabled identification systems not only to perform well on trained objects but also can report unknown objects that it would make a wrong decision. Furthermore, for dependable machine learning in practical scenarios, we need to understand how a neural network associates an input with a corresponding label. These two aspects are rarely covered in signal identification, thus motivating our research.

\section{Problem definition}
\label{sectPD}
In this research, we focus on deriving protocol-agnostic solution to identify of IoT devices from physical layer signals. The reason is that signal features directly correspond to hardware components and reveals the identities of IoT devices.

We define that an IoT device $i$ transmits specific message with corresponding baseband signal $m_i(t)$. $m_i(t)$ is modulated into:
\begin{equation}
    M_i(t) = C_i[m_i(t)]
\end{equation}{}
Where $C_i(x)$ denotes the frequency band processing chains. At receiver $j$, the received signal becomes:
\begin{equation}
    R_{ij}(t)=S_{ij}[M_i(t)]
\end{equation}{}
Where $S_{ij}$ denotes the effect of wireless channel between $i$ and $j$. This function can incorporate the effect of attenuation or additive noise. The demodulated signal is:
\begin{align}
\hat{m}_i(t)&=S^{-1}_j\{C^{-1}_j[R_{ij}(t)]\}\\\notag
&=S^{-1}_j\{C^{-1}_j[S_{ij}[C_i[m_i(t)]]]\}
\end{align}
where $C^{-1}_j(x)$ and $S^{-1}_j(x)$ are $j$‘s estimated reverse function of $C_i(x)$ and $S_{ij}$, respectively. The estimation can hardly be idealistic. Therefore, at the receiver side, $j$, the effect of such discrepancies are reflected in $\hat{m}_i(t)$ as:
\begin{align}
\hat{m}_j(t) = r_i(t) + \delta_j(t)
\end{align}
where $r_i(t)$ is directly correlated with $m_i(t)$ while the residual, $\delta_j(t)$, is utilized to recognize a wireless device. As long as $\delta_j(t)$ is uncorrelated with messages $m_i(t)$, the recognition algorithm is protocol-agnostic. Apparently, this is a classification problem, to avoid the hardship of feature engineering, we use DNN and convert IoT device recognition problem into 3 subproblems: 
\begin{enumerate}
    \item Given message-related baseband signals from various wireless transmitters, how to extract message-independent components to develop a classifier using DNNs?
    \item How to enable our classifier to properly respond to unseen signals?
    \item How can we evaluate the distinguisability between different devices?
\end{enumerate}{}

\section{Proposed Framework}
\label{sectMM}
In this section, we first present the feature extraction methods and then introduce the zero-bias deep learning framework for accurate and interpretable identification of IoT devices. 
\subsection{Baseband demodulation}
In this research, we use an independent Software-Defined Radio (SDR) receivers, denoted as $j'$, to collect baseband signals from wireless transmitters, denoted as $\hat{m}_{j'}(t)$. Given input signal $x$, the quadrature demodulation function is defined as:
\begin{align}
\label{eqSDRDemod}
C^{-1}_{j'}(x)&=I(t)+ \boldsymbol{i} \cdot Q(t)\\\notag
&=LPF[x\cdot cos(\omega_ct+\phi_0)+\boldsymbol{i}\cdot x\cdot sin(\omega_ct+\phi_0)]
\end{align}{}
where $I(t)$ and $Q(t)$ are In-Phase and Quadrature components, respectively. $\omega_c$ and $\phi_0$ are the center frequency and the phase offset of the receiver ($j'$), respectively. $\boldsymbol{i}$ denotes the imaginary part of complex function. With  Phase Lock Loop (PLL), $\omega_c$ and $\phi_0$ are supposed to be sufficiently close to RF characteristics of device $i$. $LPF$ denotes a low-pass filter. Therefore, at $j'$, demodulated baseband is:
\begin{align}
    \hat{m}_{j'}(t) & = C^{-1}_{j'}[R_{ij'}(t)]
\end{align}{}
$\hat{m}_{j'}(t)$ is complex-valued, and its instantaneous amplitude, phase and frequency are $ ||\hat{m}_{j'}(t)||=\sqrt{I^2(t)+Q^2(t)}$, $\angle \hat{m}_{j'}(t)= tan^{-1}(\dfrac{Q(t)}{I(t)})$ and $\hat{\Omega}_{j'}(t)=\dfrac{d \angle \hat{m}_{j'}(t)}{dt}$, respectively. 

Please be noted that discrepancies exist between $\hat{m}_{j}(t)$ and $\hat{m}_{j'}(t)$. Even if the wireless channel effect at receiver $j$ and $j'$ are different, we assume that an SDR receiver could still capture the effect of each wireless device's frequency band processing chain, $C_i(x)$, to recognize them.

\subsection{Feature extraction}

For protocol-agnostic device recognition, we need to remove message-correlated part $r_i(t)$ from $\hat{m}_{j'}(t)$. In this way, we ensure that our device recognition mechanism is protocol-agnostic. In addition, we only use the first 1,024 samples of $\hat{m}_{j'}(t)$.

\subsubsection{Pesudo Noise Extraction}
\begin{figure}[]
\centering  
\subfloat[Noise extraction on typical signals.]
{%
    \includegraphics[width=0.7\linewidth]{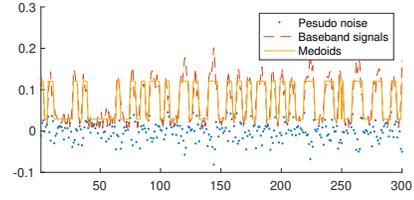}
    \label{figNoiseExtractionn}
}\\
\subfloat[Correlation coefficients of pesudo noise]
{
    \includegraphics[width=0.7\linewidth]{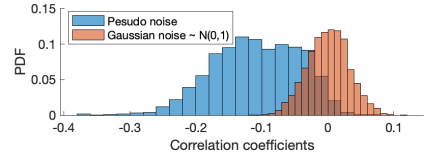}
    \label{figCorrDistrib}
}
\caption{Property of pesudo noise extraction}
\end{figure}

Suppose we have derived the numerical sequence of instantaneous metrics (amplitude, phase, or frequency), corresponding procedures are as follow:
\begin{enumerate}[\textbf{Step} 1:]
    \item We separate the sequence (denoted as $s_{j'}(n)$) into several non-overlap segments, with each segment's duration less than one symbol duration.
    \item For each segment, we perform \textit{k-medoids} algorithm on signals instantaneous phase or amplitudes with $k=2$. In essence, we use a clustering algorithm to associate numeric values to their closest medoids (representative values). Notably, we could only expect one or two possible choices of amplitudes or phases.
    \item In each segment, we generate the pesudo-noise as:
    \begin{align}
        n_{j'}(n) = s_{j'}(n) - m_k[s_{j'}(n)]
    \end{align}{}
    Where $m_k$ denotes the medoid of $s_{j'}(n)$, We subtract rationale signals from the demodulated baseband signals directly.
\end{enumerate}
A brief comparison of related signals is in Figure~\ref{figNoiseExtractionn}. Medoids could be regarded as a less noisy version of demodulated baseband signals $\hat{m}_{j'}(t)$.

The distribution of correlation coefficients (derive from 10,000 samples) of pseudo-noise against corresponding baseband signals is depicted in Figure~\ref{figCorrDistrib}. The pseudo-noise signals are weakly correlated with original messages.

\begin{figure*}[b]
\centering
\includegraphics[width=0.7\linewidth]{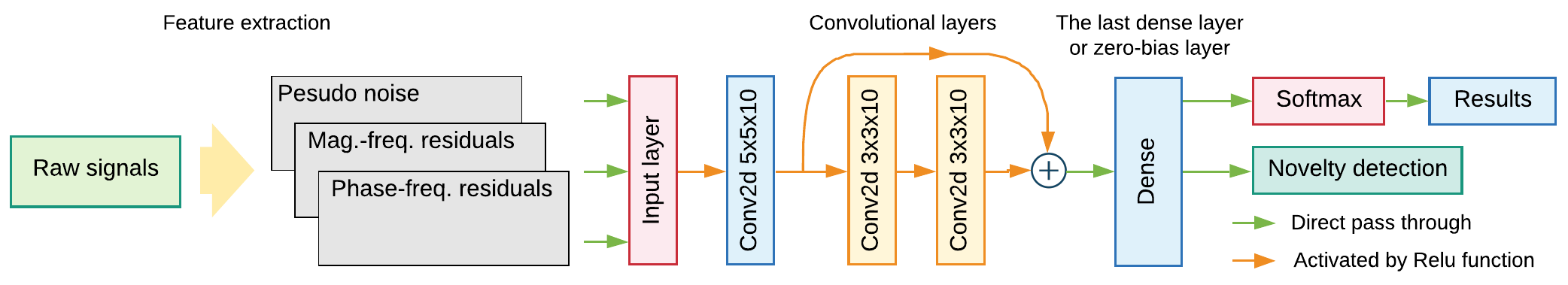}
\caption{Deep neural architecture for wireless transmitter identification.}
\label{figLearningArch}
\end{figure*}

\subsubsection{Frequency domain features}
We subtract the Fourier Transforms of both complex-valued baseband signals $\hat{m}_{j'}(t)$ and the reconstructed rationale baseband signals to extract message uncorrelated residual components in the frequency domain, formulated as:
\begin{align}
    \delta_j(\omega) = FFT[\hat{m}_{j'}(t)]-FFT[r_{j'}(t)]
\end{align}{}
where $r_{j'}(t)$ is the reconstructed rational baseband signal. Please be noted that $\hat{m}_{j'}(t)$ is complex-valued (QPSK) while $r_{j'}(t)$ can be real-valued (2FSK, 2PSK and etc.). We convert residual components into a magnitude sequence ($||\delta_j(\omega)||$), namely Mag.-Freq. residuals, and a phase sequence ($\angle \delta_j(\omega)$), namely Phase-Freq. residuals, respectively.

\subsection{Zero-Bias Deep Learning Framework for Accurate Identification of IoT Devices}
In this subsection, we present our enhancement to conventional neural networks, which is generalizable to other neural-classification problems.

The architecture of Deep learning enabled classifier for device identification is given in Figure~\ref{figLearningArch}. Convolutional layers with skip connections are employed to extract latent features, we also use a dense layer followed by a softmax layer for final classification. However, in the last dense layer, we propose a modified approach. 

Suppose we have $m$-dimension input vectors with batch size $k$, we need to convert them into $k$ $n$-dimension outputs. A conventional dense layer would perform a linear calculation as:
\begin{align}
\label{eqRegularDense}
    \boldsymbol{Y_1} = \boldsymbol{W_1} \boldsymbol{X} + \boldsymbol{b_1}
\end{align}

\begin{figure}[]
\centering
\includegraphics[width=0.85\linewidth]{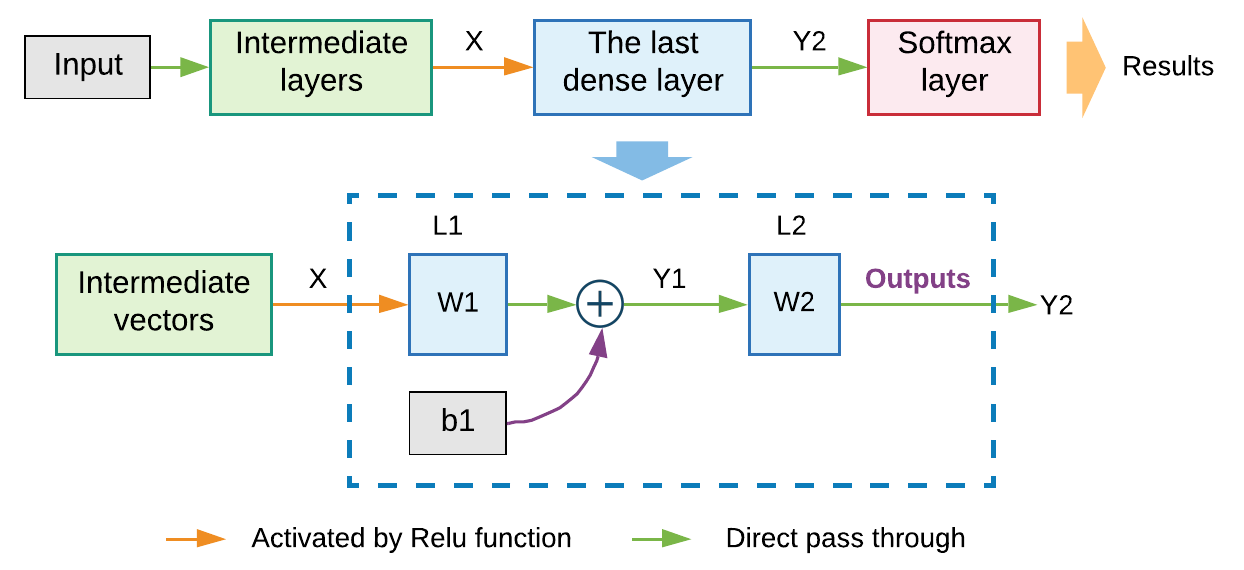}
\caption{Data flow of zero-bias dense layer.}
\label{figZerobiasDense}
\end{figure}
where $\boldsymbol{X}$, $\boldsymbol{b_1}$ and $\boldsymbol{W_1}$ denote the $m$ by $k$ input matrix, bias neurons and an $n$ by $m$ weights matrix, respectively. If we break the regular dense layer into two consecutive parts, depicted in Figure~\ref{figZerobiasDense}, a regular dense layer denoted by $L_1$ and a dense layer $L_2$ without bias, respectively. Then, Equation (\ref{eqRegularDense}) becomes:
\begin{align}
\label{eqModifiedDense}
    \boldsymbol{Y_2} =\boldsymbol{W_2}\boldsymbol{Y_1}= \boldsymbol{W_2}\boldsymbol{W_1} \boldsymbol{X} + \boldsymbol{W_2}\boldsymbol{b_1}    
\end{align}{}
Where $\boldsymbol{W_1}$ and $\boldsymbol{b_1}$ belong to $L_1$ and $\boldsymbol{W_2}$ belongs to $L_2$, respectively. Note that Equation (\ref{eqModifiedDense}) and (\ref{eqRegularDense}) are performing equivalent transforms to $\boldsymbol{X}$ and should not degrade the network performance. Moreover, in $L_2$, we can rewrite the matrix calculation into vectors:
\begin{align}
\label{eqCorrelation}
    \boldsymbol{Y_2}[\boldsymbol{y_{1k}}]=[\boldsymbol{w_{21}}\cdot\boldsymbol{y_{1k}},\boldsymbol{w_{22}}\cdot\boldsymbol{y_{1k}},...,\boldsymbol{w_{2n}}\cdot\boldsymbol{y_{1k}}]
\end{align}{}
Where $\boldsymbol{w_{21}},...,\boldsymbol{w_{2n}}$ are row vectors corresponding to $n$ output classes, $\boldsymbol{y_{1k}}$ is one of the $k$ column vectors in batch, and $\boldsymbol{Y_2}[\boldsymbol{y_{1k}}]$ is the output vector. The process in equation (\ref{eqCorrelation}) can be rewritten using \textit{Cosine Similarity}:
\begin{align}
\label{eqCosineSim}
    \boldsymbol{w_{2n}}\cdot\boldsymbol{y_{1k}} = ||\boldsymbol{w_{2n}}||\cdot||\boldsymbol{y_{1k}}||\cdot cos(\boldsymbol{w_{2n}},\boldsymbol{y_{1k}})
\end{align}{}
If $L_2$ is followed by a Softmax layer and we take $\boldsymbol{w_{21}},...,\boldsymbol{w_{2n}}$ as fingerprints of classes $1$ to $n$, we conclude that $L_2$ actually calculates a scaled version of cosine similarities among input against fingerprints of target classes. 

Moreover, we can safely generalize this discovery to understand the behavior of last dense layers in neural networks:

\begin{remark}[Property of dense layers]
\label{rmConfidenceFactors}
If an output vector of a dense layer represent the degrees of confidence of corresponding class/position against an input, then each confidence degree is jointly controlled by the magnitude of the class/position-related fingerprint, the fingerprint's cosine similarity to the input, and the bias neuron of this class. 
\end{remark}{}
Although the magnitude of an input feature vector $||\boldsymbol{y_{1k}}||$ seems to take effect as in Equation (\ref{eqCosineSim}), but in the consecutive Softmax layer, the magnitude $||\boldsymbol{y_{1k}}||$ only contributes to a common base number as in Equation (\ref{eqSoftmax}): 
\begin{align}
\label{eqSoftmax}
    class=\dfrac{\exp[{||\boldsymbol{y_{1k}}||}\cdot ||\boldsymbol{w_{2n}}|| \cdot cos(\boldsymbol{w_{2n}},\boldsymbol{y_{1k}})] }{\sum_n \exp[{||\boldsymbol{y_{1k}}||}\cdot ||\boldsymbol{w_{2n}}|| \cdot cos(\boldsymbol{w_{2n}},\boldsymbol{y_{1k}})]}
\end{align}{}
Where the base number, $\exp{||\boldsymbol{y_{1k}}||}$ only controls the steepness of the monotonic mapping curve 
According to Remark~\ref{rmConfidenceFactors}, we can derive another important remark:
\begin{remark}[Neural networks' partiality]
As long as prior layers do not converge to constant functions, A neural network's partiality to specific classes is encoded in its last dense layer before Softmax, and the bias is jointly controlled by the magnitude of class-related fingerprint vector and the bias neuron of the corresponding class. 
\end{remark}{}

In our proposed paradigm of dense layer without bias neurons, we can derive more specific corollaries:
\begin{corollary}[Fingerprints' magnitude]
\label{rmVarianceFingerprintMag}
If the variance of the magnitude of fingerprints vectors is small, the layer $L_2$ has less bias to specific classes.
\end{corollary}{}
Currently, we have two approaches to remove the unwanted effects of fingerprint vectors' magnitudes:
\begin{itemize}
    \item We can use regularization to eliminate the variance of fingerprints, we make their values relative close;
    \item We can replace Equation (\ref{eqCorrelation}) with Equation (\ref{eqUnitFingerprint}):
    \begin{align}
    \label{eqUnitFingerprint}
        &\boldsymbol{Y_2} = [\dfrac{\boldsymbol{w_{21}}}{\sqrt{\boldsymbol{w^2_{21}}}},...,\dfrac{\boldsymbol{w_{2n}}}{\sqrt{\boldsymbol{w^2_{2n}}}}]^T[\boldsymbol{y_{11}},...,\boldsymbol{y_{1k}}]      
    \end{align}
    Moreover, we can eliminate the side effects of feature vectors' magnitude at the same time:
    \begin{align}
    \label{eqUnitFingerprint2}
        \boldsymbol{Y_2} = \lambda [\dfrac{\boldsymbol{w_{21}}}{\sqrt{\boldsymbol{w^2_{21}}}},...,\dfrac{\boldsymbol{w_{2n}}}{\sqrt{\boldsymbol{w^2_{2n}}}}]^T
        [\dfrac{\boldsymbol{y_{11}}}{\sqrt{\boldsymbol{y^2_{11}}}},...,\dfrac{\boldsymbol{y_{1k}}}{\sqrt{\boldsymbol{y^2_{1k}}}} ]
    \end{align}  
    Where $\lambda$ is a trainable value to provide the freedom of controlling the steepness of the mapping curve in the Softmax layer. Please be noted that $\boldsymbol{Y_2}$s are differentiable in these two scenarios and Equation (\ref{eqUnitFingerprint}) is still equivalent to linear operations.
\end{itemize}{}

We eliminate the classifiers' partiality or bias. We treat the possibility of each class equally and it's the essence of "zero-bias" dense layer. With the zero-bias enhancement, we have corollary \ref{rmFingerprintMutualInfo}: 
\begin{corollary}[Fingerprints' mutual distances]
\label{rmFingerprintMutualInfo}
Fingerprints in the zero bias dense layer ($L_2$) act as angular representatives of corresponding classes and should has sufficiently small mutual cosine similarities.
\end{corollary}{}

\begin{figure}[]
\centering
\includegraphics[width=0.65\linewidth]{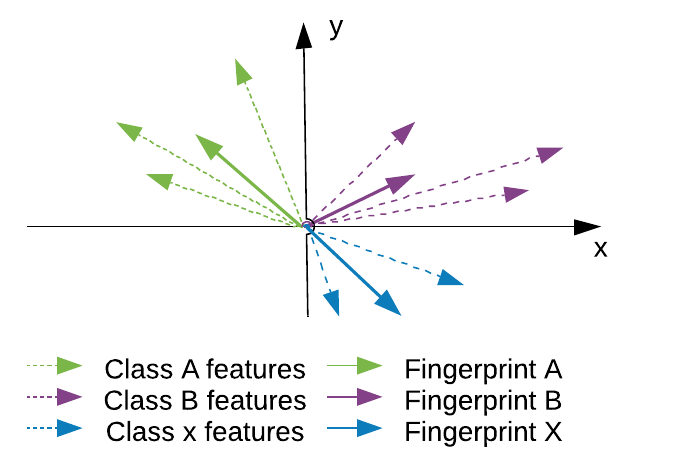}
\caption{Relation of fingerprint vectors and feature vectors.}
\label{figFingerprintVectors}
\end{figure}

A simplified example of corollary \ref{rmFingerprintMutualInfo} is given in Figure~\ref{figFingerprintVectors}, suppose we have three classes (A, B, and X) for a deep neural network to distinguish from, the fingerprint vector of each class only captures a representative direction. With this property, we only need to insert or remove fingerprints in $L_2$, to register or remove corresponding classes. 

\begin{figure}
\centering  
\subfloat[After 1 epoch]
{%
    \includegraphics[width=0.45\linewidth]{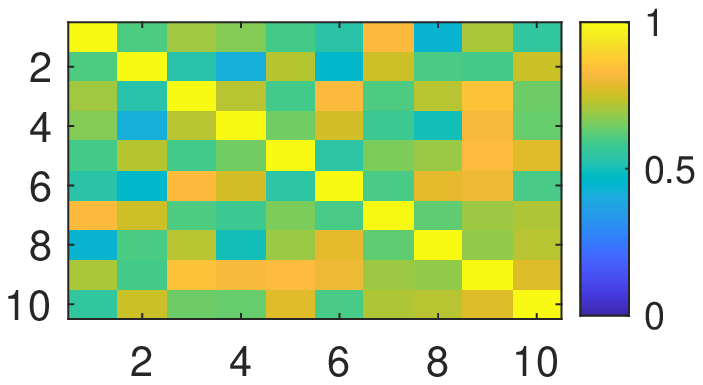}
    \label{figBadlyTrained}
}
\subfloat[After 10 epoches]
{%
    \includegraphics[width=0.45\linewidth]{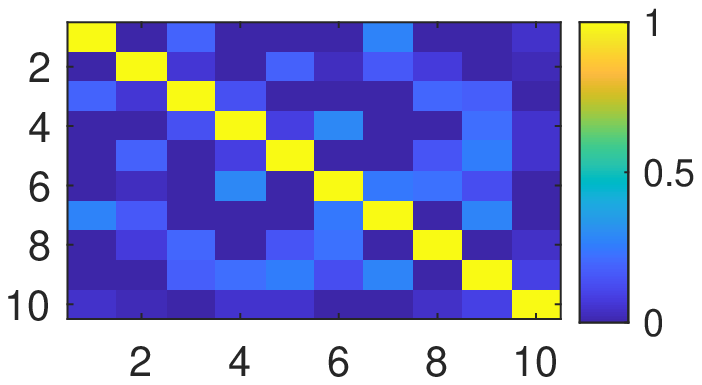}
    \label{figWellTrained}
}
\caption{Fingerprint distance matrix of Minst example}
\label{figFDM}
\end{figure}

Another benefit is to evaluate how well different classes are mutually distinguishable from each other. We can construct a Fingerprint Distance (FD) matrix as:
\begin{align}
    FD = \begin{bmatrix} cos(\boldsymbol{w_1},\boldsymbol{w_1})&\dots &cos(\boldsymbol{w_1},\boldsymbol{w_n})\\\vdots & \ddots & \vdots \\
    cos(\boldsymbol{w_n},\boldsymbol{w_1})&\dots &cos(\boldsymbol{w_n},\boldsymbol{w_n})
    \end{bmatrix}
\end{align}
This matrix can directly reflect how well different classes are separated in the latent space. We replace the last dense layer with zero-bias dense layer (contains both $L_1$ and $L_2$) in the MNIST example \cite{matlabMinst} and plot the FD matrices when training accuracy reaches 60.2\% and 95.8\%, respectively. As in Figure~\ref{figFDM}, fingerprints are distantly separated with higher accuracy. 

In this subsection, we propose a new scheme of creating zero-bias neural networks and a thorough analysis of the mechanism of dense layers. A summary of our the enhancement is:
\begin{remark}[Zero-bias layer enhancement]
We replace the last dense layer of a neural network with a consecutive structure consisting of a regular dense layer ($L_1$) and a zero-bias similarity comparing layer ($L_2$).
\end{remark}
We notice that some researches directly employ Equation (\ref{eqUnitFingerprint2}) as cosine similarity \cite{gidaris2018dynamic,luo2018cosine} in deep learning, we differentiate from them as: a) we provided a mathematically equivalent transform, by using another regular fully connected layer $L_1$. b) our experiments show that directly applying cosine similarity without $L_1$ dramatically increases the difficulty of training.

\subsection{Novel device identification}
A wireless device identification system needs to identify anomalous signals from novel devices. In a conventional neural network, the Softmax layer associates labels to the largest activation. Such behavior would result in wrong answers given falsified signals from unknown devices. Suppose that the zero-bias layer enhancement in Equation (\ref{eqUnitFingerprint2}) is applied, the output of the layer directly represent cosine similarities. We define the concept \textit{Similarity Response} as:
\begin{definition}[Similarity response]
For an input, the maximum value in output vector after zero-bias or regular dense layer is defined as its similarity response.
\end{definition}
An unknown device with false identity can be detected if its signals' similarity reponses are below a reasonable threshold. For example, if the similarity response of known devices follows a Gaussian distribution, $N(m_k,\sigma_k)$, an input with the highest similarity less than $m_k - \sigma_k$ can be subject to novel or even spoofing device. 

\section{Performance Evaluation}
Automatic Dependent Surveillance-Broadcast (ADS-B) \cite{riddle1090}, which accurately observe and track air traffic, is a fundamental safety infrastructure modern aviation. This system is designed to be simple and widely adaptable but it's extremely vulnerable to identity spoofing attacks. In this section, we present our performance evaluation results using real ADS-B data and demonstrate how our proposal could be elegantly applied in practical systems.
\label{sectEED}
\subsection{Evaluation dataset}
\begin{figure}[]
\centering
\includegraphics[width=0.8\linewidth]{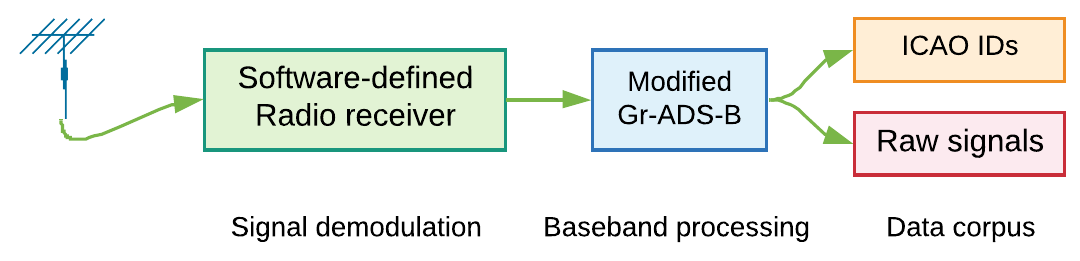}
\caption{Collection of ADS-B signals.}
\label{figDataCollectionFlow}
\end{figure}

\begin{figure}[]
\centering
\includegraphics[width=0.75\linewidth]{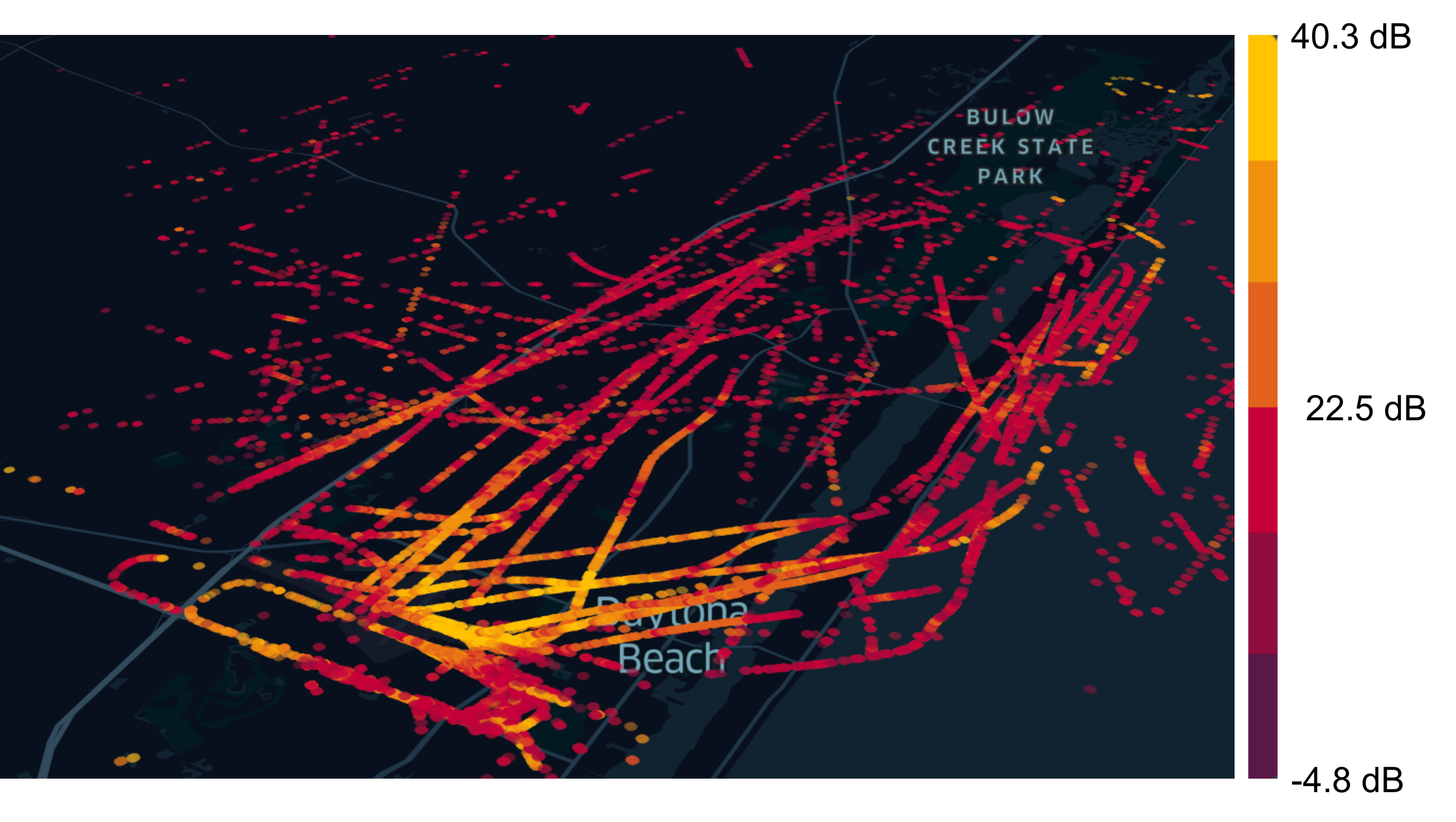}
\caption{Geographic distribution of aircraft transponders.}
\label{figADSBGeoHeatmap}
\end{figure}

Nowadays, Commercial aircraft are equipped with dedicate 1090MHz transponders to broadcast its geo-coordinates, velocities, altitudes, headings as well as their unique identifiers, a.k.a International Civil Aviation Organization (ICAO) IDs. Such signals provides a great variety of signals from known wireless devices. In our data collection pipeline depicted in Figure~\ref{figDataCollectionFlow}, we used a modified \textit{gr-adsb} library to decode ADS-B messages and store raw baseband digital signals. We collected the ADS-B signal from more than 140 aircraft at Daytona Beach international airport (ICAO: DAB) for 24 hours (Jan $4^{th}$, 2020) using a Software-Defined Radio receiver (USRP B210). The receiver is configured with a sample rate of 8 MHz. During this period, more than 30,000 ADS-B messages are collected with coordinates and SNR (in colors) depicted in Figure~\ref{figADSBGeoHeatmap}.

\subsection{Known device verification}
\begin{figure}[ht]
\centering
\includegraphics[width=0.9\linewidth]{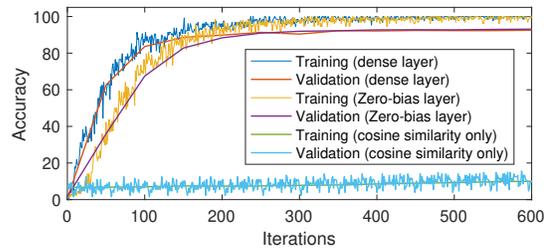}
\caption{Comparison of training performance.}
\label{figCompareTraining}
\end{figure}

We first conduct a general performance test of the system (depicted in Figure~\ref{figLearningArch}). As depicted, the deep learning model can associate received signals with accuracy greater than 94.3\%. Furthermore, a brief comparison of DNN with proposed zero-bias layer, regular dense layer and only cosine similarity before softmax\footnote{Similar network architecture with cosine similarity and softmax directly after convolution filters.} on the same dataset is given in Figure~\ref{figCompareTraining}. As depicted, DNNs with zero-bias layer or regular dense layer reach almost identical performance. However, the zero-bias layer requires more training iterations, and its rising rate of accuracy is lower at the beginning. Interestingly, if we only use cosine similarity directly after convolutions, the deep learning system can not converge.

\begin{figure}[]
\centering
\includegraphics[width=0.7\linewidth]{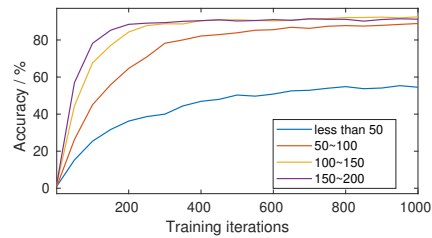}
\caption{Validation accuracy in terms of training data size for each transmitter.}
\label{figValidationVsDataQuantity}
\end{figure}

To evaluate the deep learning model in terms of training data quantity, we manually limit the number of samples of each transmitter in the training set and use this specially "reduced" training set to train the zero-bias DNN model. As depicted in Figure~\ref{figValidationVsDataQuantity}, the model converges after 800 iterations (40 epochs) and show that we only need 200 samples to recognize each transmitter.

\subsection{Novel Device Identification}
We randomly pick ADS-B signals from 30 aircraft to train the neural network and use signals from the remaining 120 aircraft as unseen novel devices' signals. We compare the performance of our zero-bias layer, regular dense layer, and one-class Support Vector Machine (SVM), respectively. In this subsection we define the optimal decision boundary as:
\begin{align}
    max_{\tau} ||cdf(P_u(\tau)) - cdf(P_k(\tau))||
\end{align}
where $P_u(\tau)$ and $P_k(\tau)$ are probability distribution functions of similarity response of unknown and known devices. $cdf(\cdot)$ denotes the cumulative density function.
\subsubsection{Zero-bias and regular dense layer}
\begin{figure*}[]
\centering  
\subfloat[Zero-bias DNN]
{%
    \includegraphics[width=0.35\linewidth]{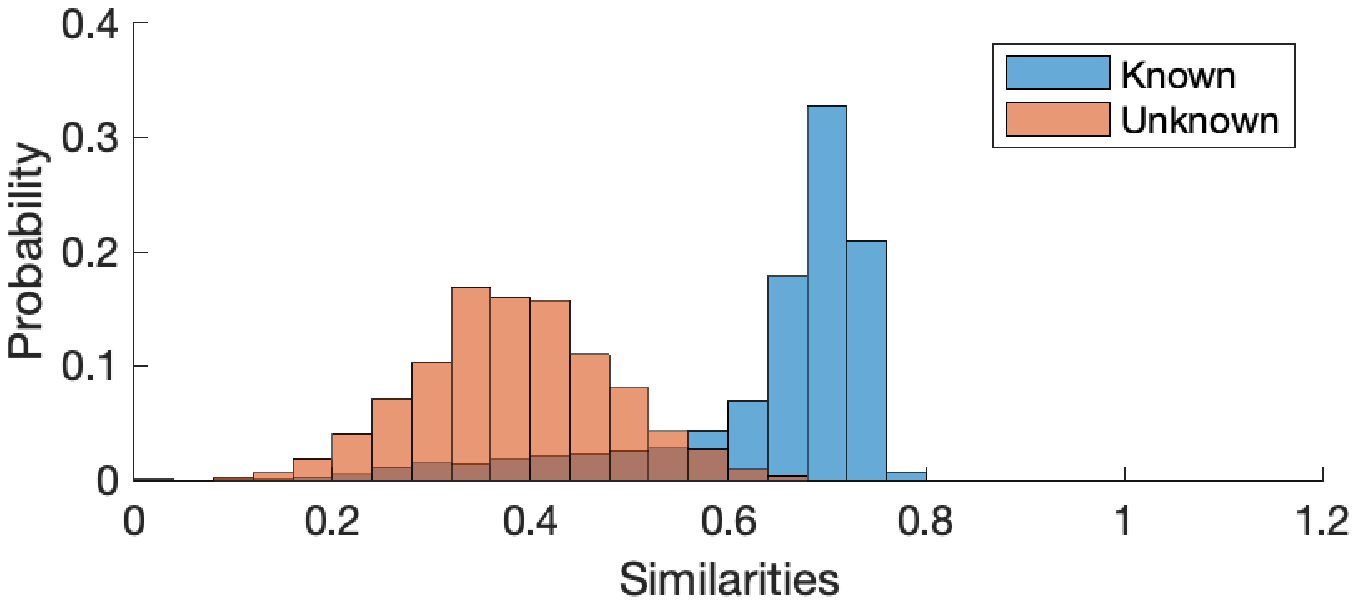}
    \label{figPdfFingerprint}
}\hspace*{-1.5em}
\subfloat[Regular DNN]
{%
    \includegraphics[width=0.35\linewidth]{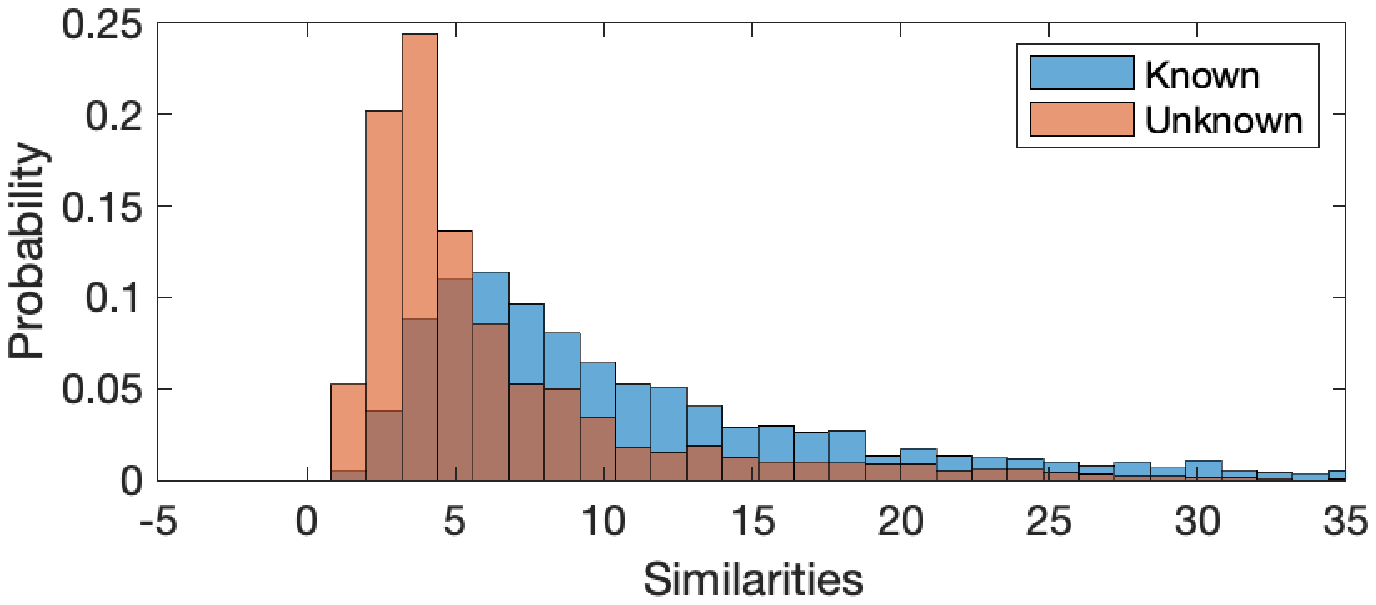}
    \label{figDensePdfSimilarities}
}\hspace*{-1.5em}
\subfloat[One-class SVM]
{%
    \includegraphics[width=0.35\linewidth]{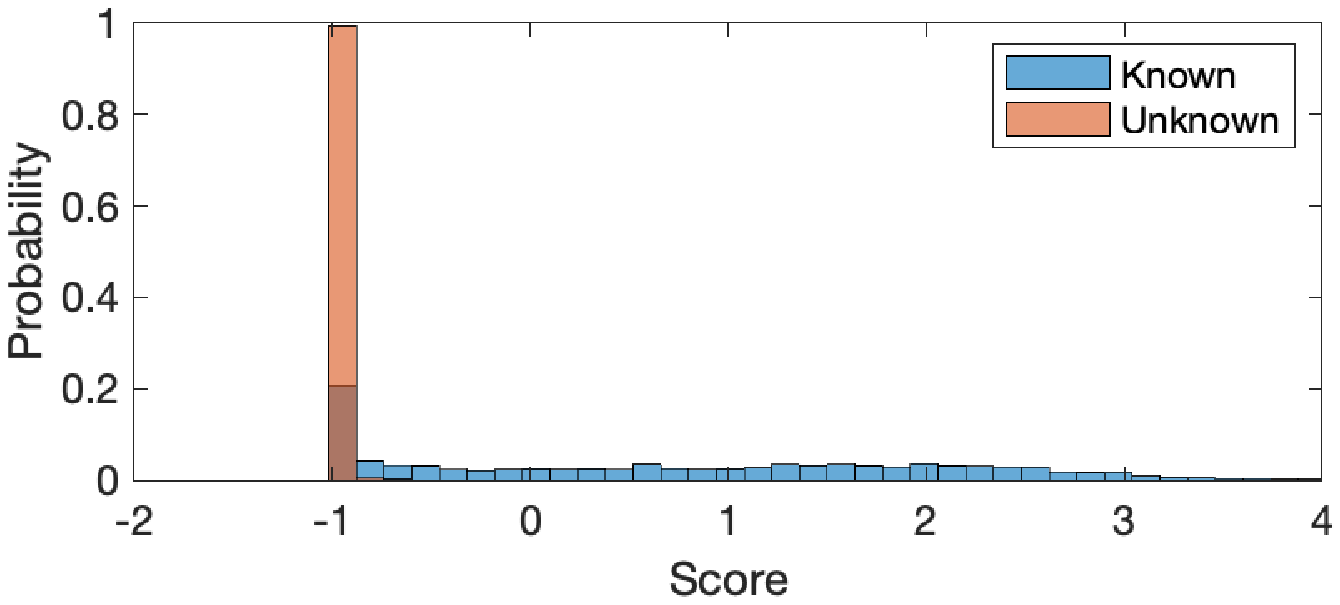}
    \label{figPdfOneClass}
}
\\
\subfloat[Zero-bias DNN]
{
    \includegraphics[width=0.35\linewidth]{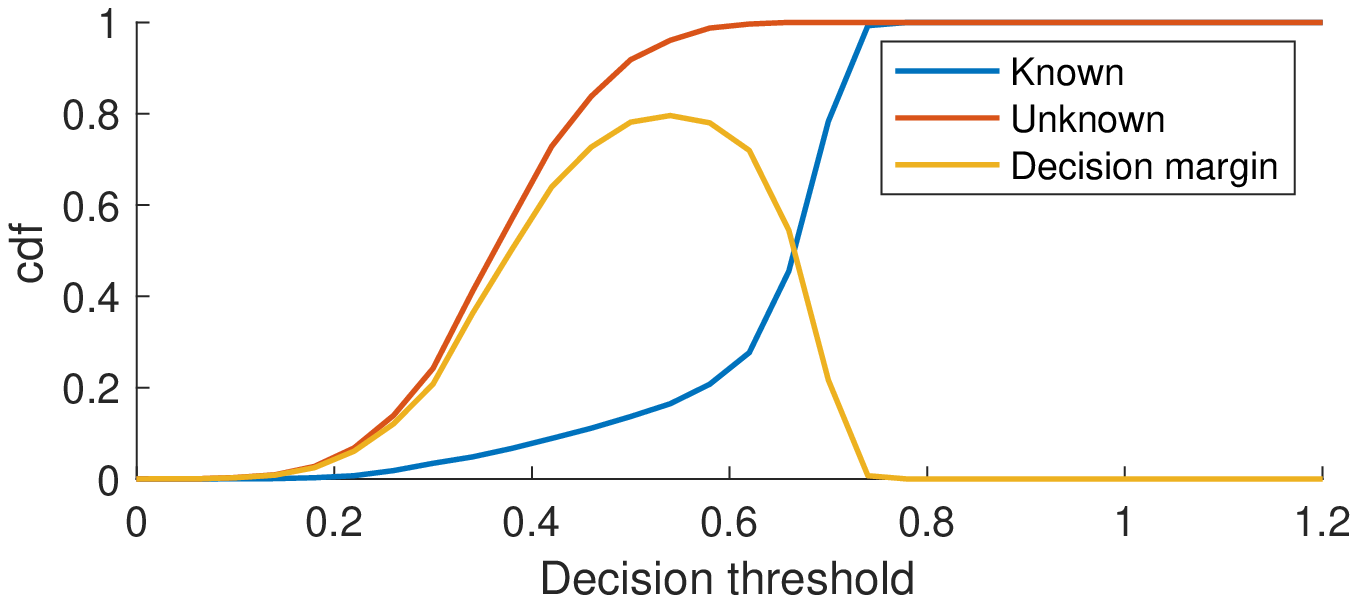}
    \label{figCdfFingerprint2}
}\hspace*{-2.0em}
\subfloat[Regular DNN]
{
    \includegraphics[width=0.35\linewidth]{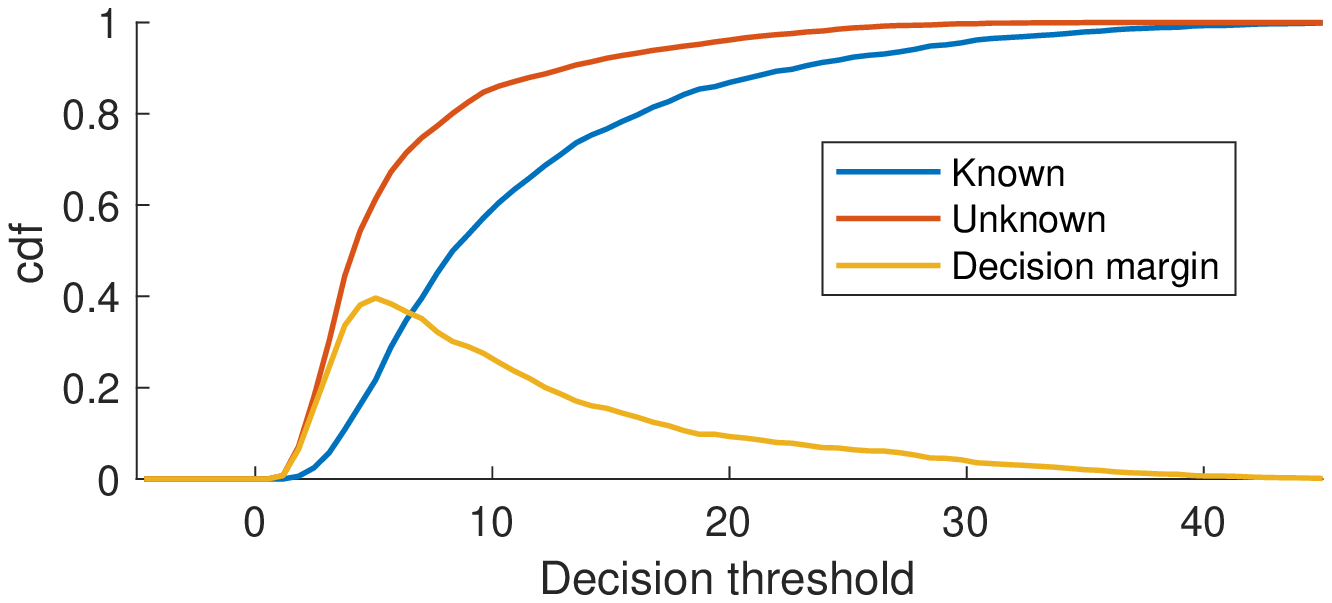}
    \label{figDenseCdfSimilarities}
}\hspace*{-2.0em}
\subfloat[One-class SVM]
{
    \includegraphics[width=0.35\linewidth]{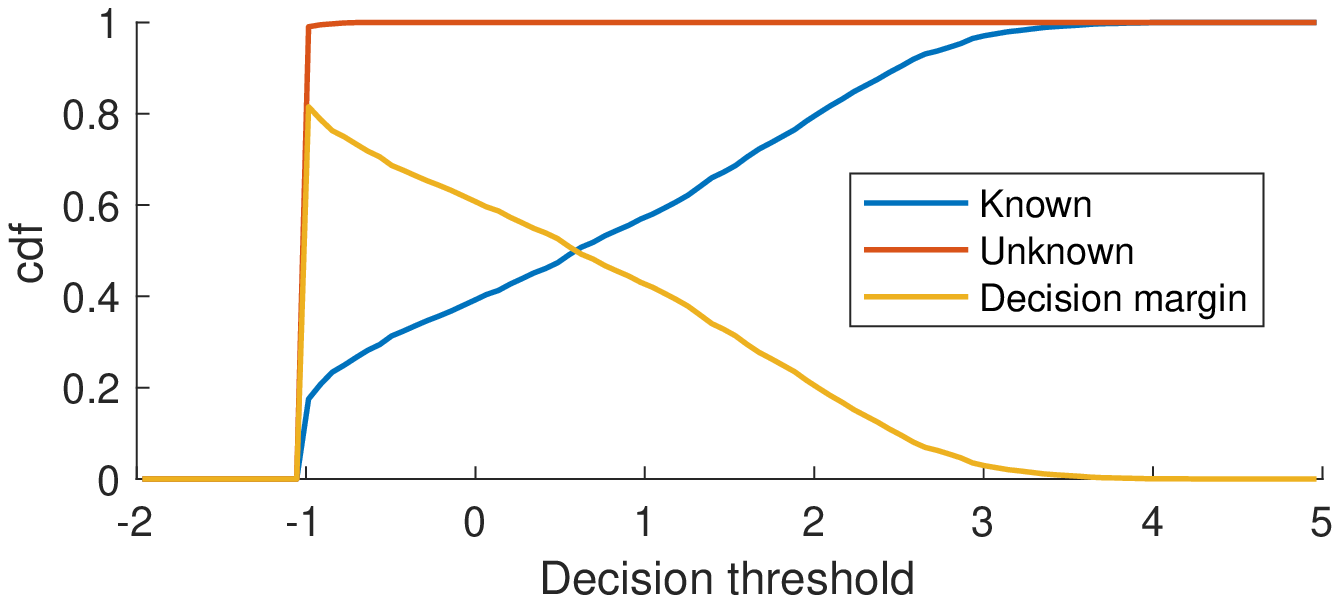}
    \label{figCdfOneClass}
}
\caption{Performance of Threshold based anomaly detections}
\end{figure*}

We employ the zero-bias layer (use Equation (\ref{eqUnitFingerprint2})) for final output. The probability distribution and decision thresholds are given in Figure~\ref{figPdfFingerprint} and \ref{figCdfFingerprint2}, respectively. Figure~\ref{figPdfFingerprint} demonstrates that the similarities response of unknown signals are higher than unknown signals in most cases. Figure~\ref{figCdfFingerprint2} shows that we can easily select an optimum separation threshold to maximize the decision boundary of the anomaly detection algorithm. In our application, we choose the median value of similarity responses on known signals minus its standard deviation as a decision threshold. 

We train the an identical neural network but with the zero-bias layer replaced by regular dense layer. But the anomaly detection performances are much worse, as depicted in Figure~\ref{figDensePdfSimilarities} and \ref{figDenseCdfSimilarities}, the similarity response of regular dense layer on known and unknown data are severely overlapped. The optimal decision boundary in this scenario is small.

\subsubsection{One-class SVM}
We use the feature vectors in training signals (directly produce by convolutional layers) of zero-bias DNN to train a one-class SVM model, we then use feature vectors from validation set as unseen signals to test the performance of one-class SVM. We collect the prediction scores on both known signals and unknown signals with statistic results presented in Figure~\ref{figPdfOneClass} and \ref{figCdfOneClass}, respectively. The result indicates that the prediction scores of known devices' signal occupy a much wider area (larger variance), which may cause difficulty for choosing the right threshold. The fact indicates that performance of the zero-bias layer enabled DNN in anomaly detection is comparable with one-class SVM. However, in our experiment, the one-class SVM model ultimately stores more than 5,000 support vectors, while the zero-bias layer only stores directional fingerprints of known aircraft transponders (less than 200). Therefore, we believe our solution is more adaptable for real-time machine learning.

\section{Conclusion}
\label{sectCC}
In this paper, we propose a novel deep learning framework for IoT device identification. Different from existing works, we focus on how to enable deep learning to be practically usable and dependable. Our contributions are as follows: Firstly, we analyze the mathematical essence of IoT device identification and use residual signals to identify real-world ADS-B transmitters. We got a promising recognition rate of 94\% among more than 130 airborne transponders. Secondly, we thoroughly analyze the behavior of the last fully-connected layer in deep neural networks and propose our improvement, the zero-bias layer, for interpretable and dependable machine learning in IoT. Experiments show that we obtain equivalent accuracy compared to the regular deep neural network, but obtain much better performances in terms of anomaly detection. Therefore, we believe the zero-bias layer can be generalized to other domains, such as virus detection or unsupervise intrusion detection. In the future, we will focus on how to efficiently discover reusable function blocks in pre-trained networks and apply them to new domains.

\section*{Acknowledgment}

This research was partially supported through Embry-Riddle Aeronautical University's Faculty Innovative Research in Science and Technology (FIRST) Program and the National Science Foundation under grant No. 1956193.




\bibliographystyle{IEEEtran}

\bibliography{ReviewRef.bib}
%

%
\vspace{-2em}
\begin{IEEEbiography}[{\includegraphics[width=1in,height=1.25in,clip,keepaspectratio]{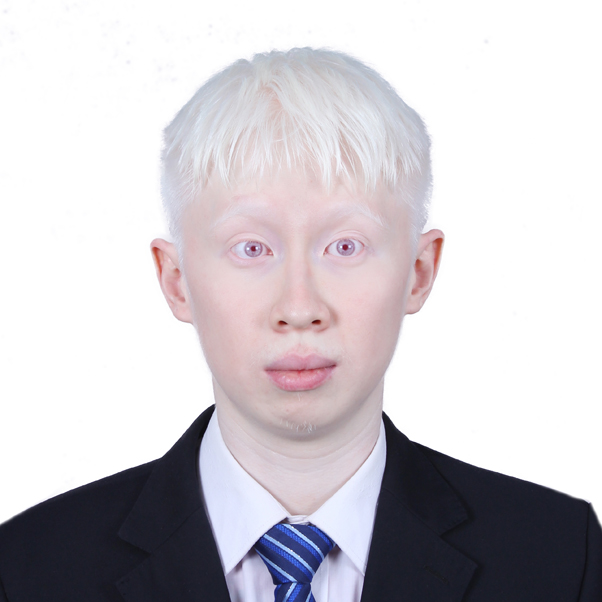}}]{YongXin Liu}
(LIU11@my.erau.edu) received his first Ph.D. from South China University of Technology (SCUT) and currently working towards his second Ph.D. in the Department of Electrical Engineering and Computer Science, Embry-Riddle Aeronautical University, Daytona Beach, FL. His major research interests include data mining, wireless networks, the Internet of Things, and unmanned aerial vehicles. 
\end{IEEEbiography}
\vspace{-6em}
\begin{IEEEbiography}[{\includegraphics[width=1in,height=1.25in,clip,keepaspectratio]{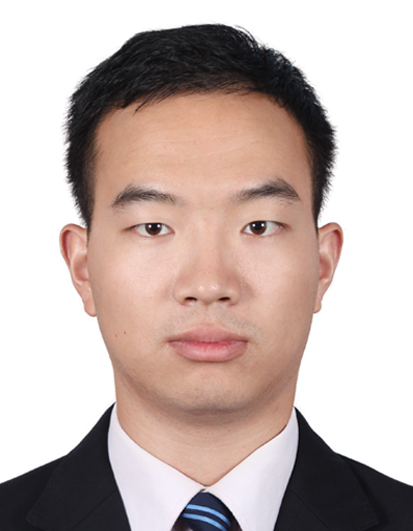}}]{Jian Wang}
(wangj14@my.erau.edu) is a Ph.D. student in the Department of Electrical Engineering and Computer Science, Embry-Riddle Aeronautical University (ERAU), Daytona Beach, Florida, and a graduate research assistant in the Security and Optimization for Networked Globe Laboratory (SONG Lab, www.SONGLab.us). He received his M.S. from South China Agricultural University (SCAU) in 2017. His research interests include wireless networks, unmanned aerial systems, and machine learning.
\end{IEEEbiography}
\vspace{-5em}
\begin{IEEEbiography}[{\includegraphics[width=1in,height=1.25in,clip,keepaspectratio]{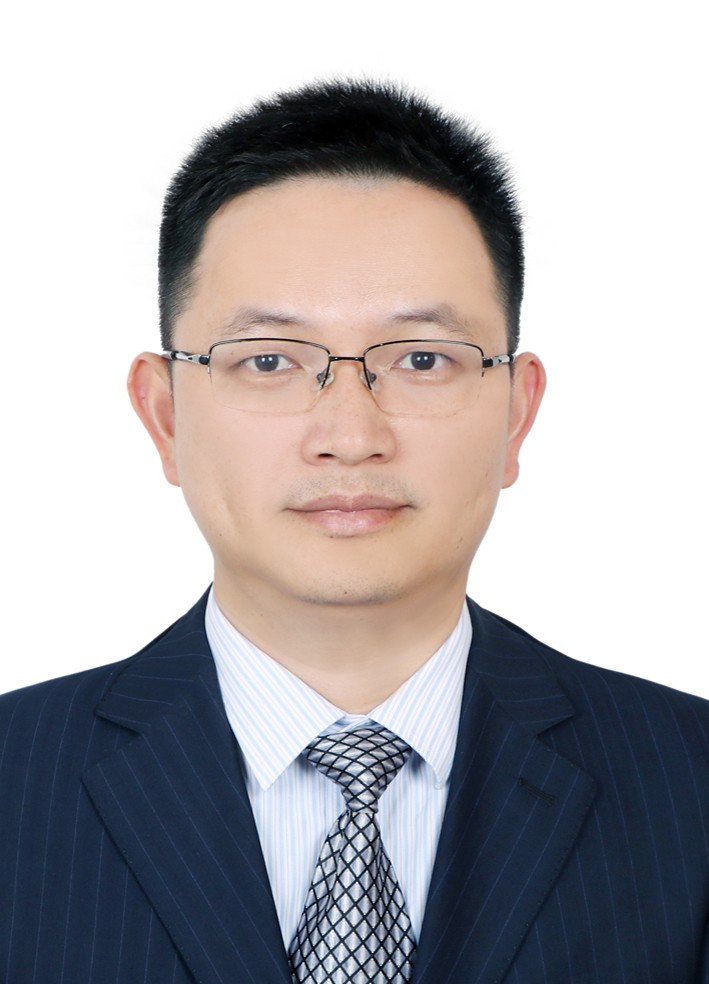}}]{Jianqiang Li}
(lijq@szu.edu.cn) received his B.S. and Ph.D.
degrees from the South China University of
Technology in 2003 and 2008, respectively. He is a Professor with the College of Computer
and Software Engineering, Shenzhen University,
Shenzhen, China. His major
research interests include Internet of Things, robotic,
hybrid systems, and embedded systems.
\end{IEEEbiography}
\vspace{-6em}
\begin{IEEEbiography}[{\includegraphics[width=1in,height=1.1in,clip,keepaspectratio]{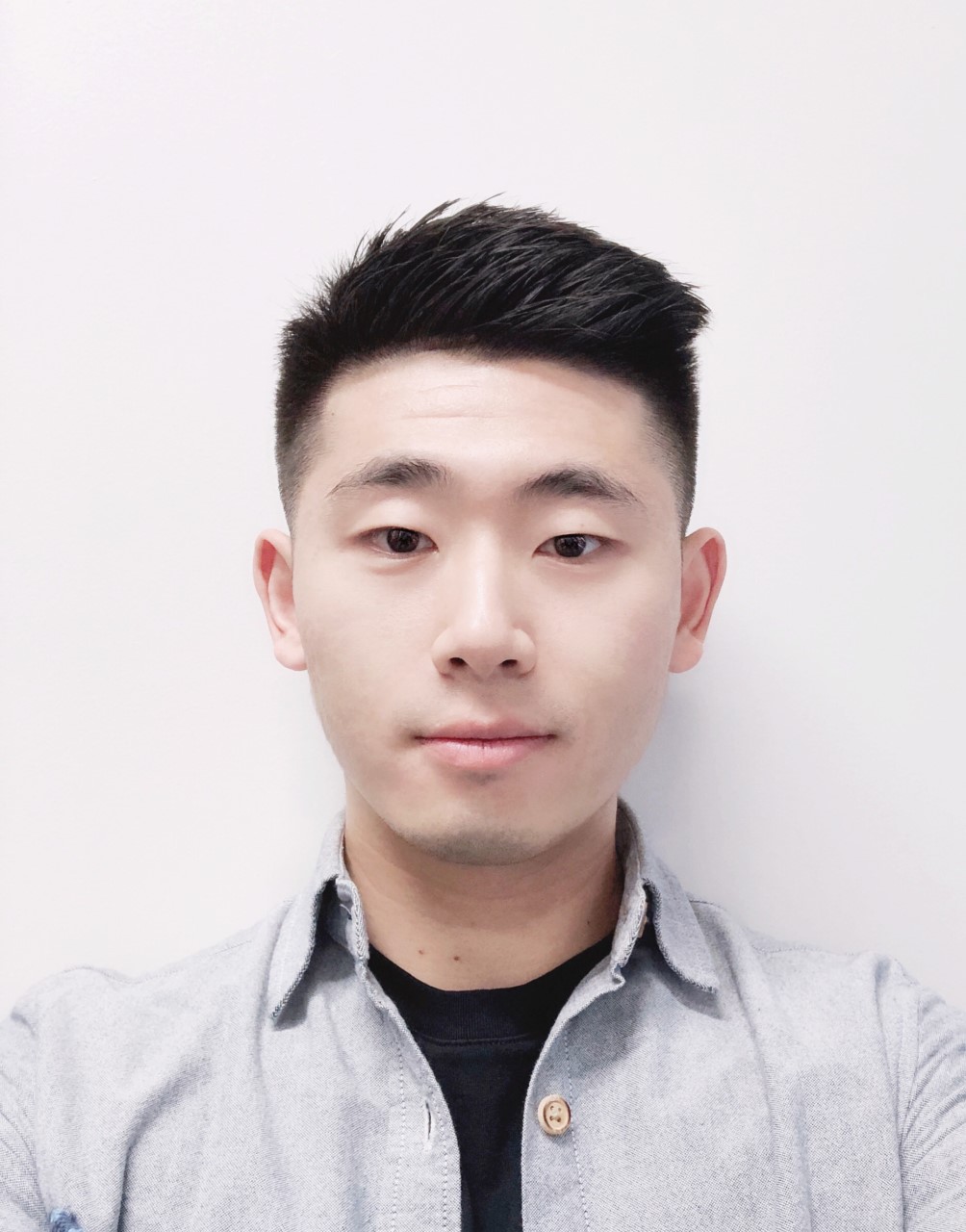}}]{Shuteng Niu}
(shutengn@my.erau.edu) is a Ph.D. student in the Department of Electrical Engineering and Computer Science, Embry-Riddle Aeronautical University (ERAU), Daytona Beach, Florida, and a graduate research assistant in the Security and Optimization for Networked Globe Laboratory (SONG Lab, www.SONGLab.us). He received his M.S. from ERAU in 2018. His research interests include machine learning, data mining, and signal processing.
\end{IEEEbiography}
\vspace{-6em}
\begin{IEEEbiography}[{\includegraphics[width=1in,height=1.25in,clip,keepaspectratio]{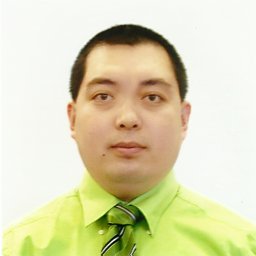}}]{Houbing Song} (M'12-SM'14) received his Ph.D. degree in electrical engineering from the University of Virginia, Charlottesville, in 2012. In August 2017, he joined the Department of Electrical Engineering and Computer Science, Embry-Riddle Aeronautical University, where he is currently an assistant professor and the Director of the Security and Optimization for Networked Globe Laboratory (SONG Lab, www.SONGLab.us). He serves as an Associate Technical Editor for IEEE Communications Magazine and an Associate Editor for IEEE Internet of Things Journal. 
\end{IEEEbiography}
\vspace{-5em}
\begin{IEEEbiography}[{\includegraphics[width=1in,height=1.25in,clip,keepaspectratio]{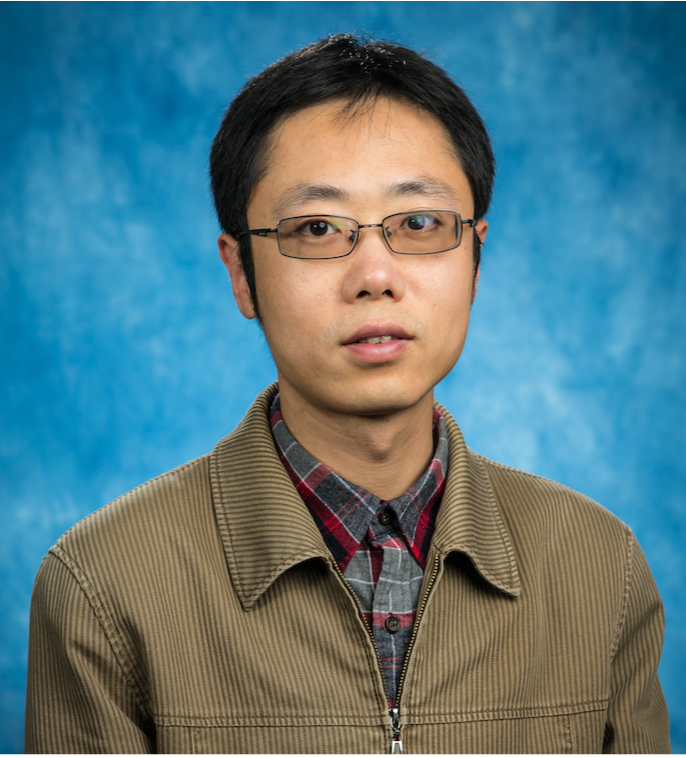}}]{Thomas Yang}
(yang482@erau.edu) received his Ph.D. in Electrical Engineering in 2004 from the University of Central Florida, Orlando, Florida. He is currently a Full Professor of Electrical and Computer Engineering at Embry-Riddle Aeronautical University, Daytona Beach, Florida. Dr. Yang's research interests include signal processing for wireless communication, autonomous multi-agent systems, and machine learning.
\end{IEEEbiography}
\vspace{-6em}
\begin{IEEEbiography}[{\includegraphics[width=1in,height=1.4in,clip]{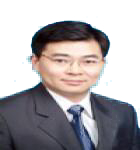}}]{Zhongming}
(mingz@szu.edu.cn) is a Professor with the College of Computer and Software Engineering, Shenzhen University. He led three projects of the National Natural Science Foundation, and two projects of the Natural Science Foundation of Guangdong Province,China. His major research interests include home networks, Internet of Things, and cloud computing. He is a Senior Member of the Chinese Computer
Federation.
\end{IEEEbiography}




\end{document}